\documentclass[conference]{IEEEtran}
\IEEEoverridecommandlockouts
\usepackage{cite}
\usepackage{amsmath,amssymb,amsfonts}
\usepackage{algorithm}
\usepackage[noend]{algpseudocode}
\usepackage{stfloats}
\makeatletter

\def\BState{\State\hskip-\ALG@thistlm}
\makeatother

\usepackage{graphicx}
\usepackage{textcomp}
\usepackage{xcolor}
\usepackage{hyperref}

\def\BibTeX{{\rm B\kern-.05em{\sc i\kern-.025em b}\kern-.08em
    T\kern-.1667em\lower.7ex\hbox{E}\kern-.125emX}}

\usepackage{caption}

\usepackage{authblk}

\begin{document}
\small
\title{Tessellated Linear Model for Age Prediction from Voice\\
{\footnotesize \textsuperscript{}}
\thanks{}
}

% Author names with affiliations
\author[1]{\textit{Dareen Alharthi}}
\author[2]{\textit{Mahsa Zamani}}
\author[1]{\textit{Bhiksha Raj}}
\author[1]{\textit{Rita Singh}}

% Affiliations
\affil[1]{Carnegie Mellon University}
\affil[2]{freelancer}

% \name{Dareen Alharthi$^1$, Mahsa Zamani$^1$, Bhiksha Raj$^{1,2}$, Rita Singh$^1$}
% \address{$^1$Carnegie Mellon University}

% \author{\IEEEauthorblockN{1\textsuperscript{st} Dareen Alharthi }
% \IEEEauthorblockA{\textit{Language Technologies Institute} \\
% \textit{Carnegie Mellon University}\\
% Pittsburgh, United States \\
% dalharth@andrew.cmu.edu}
% \and
% \IEEEauthorblockN{2\textsuperscript{nd}Mahsa Zamani
% }
% \IEEEauthorblockA{\textit{Language Technologies Institute} \\
% \textit{Carnegie Mellon University}\\
% Pittsburgh, United States \\}
% \and
% \IEEEauthorblockN{3\textsuperscript{rd} Bhiksha Raj}
% \IEEEauthorblockA{\textit{Electrical and Computer Engineering} \\
% \textit{Carnegie Mellon University}\\
% Pittsburgh, United States \\
% bhikshar@andrew.cmu.edu}

% \and
% \IEEEauthorblockN{4\textsuperscript{th} Rita Signh}
% \IEEEauthorblockA{\textit{Language Technologies Institute} \\
% \textit{Carnegie Mellon University}\\
% Pittsburgh, United States \\
% rsingh@andrew.cmu.edu}

% \and
% \and

\maketitle

\begin{abstract}
Voice biometric tasks, such as age estimation require modelling the often complex relationship between voice features and the biometric variable.
%, often exhibit complex patterns that simple models struggle to capture effectively. 
While deep learning models can handle such complexity, 
%by learning intricate relationships within the data, 
they typically require large amounts of accurately labeled data to perform well. Such data are often scarce for biometric tasks such as voice-based age prediction.
%, labeled datasets are scarce, making it challenging to train deep models effectively. 
On the other hand, simpler models like linear regression can work with smaller datasets but often fail to generalize to the underlying non-linear patterns present in the data. In this paper we propose the Tessellated Linear Model (TLM), a piecewise linear approach that combines the simplicity of linear models with the capacity of non-linear functions. TLM tessellates the feature space into convex regions and fits a linear model within each region.  We optimize the tessellation and the linear models using a hierarchical greedy partitioning.
%, which builds classifiers to define partition boundaries. 
We evaluated TLM on the TIMIT dataset on the task of age prediction from voice, where it outperformed state-of-the-art deep learning models.  The source code will be made publicly available\footnote{\href{https://github.com/DareenHarthi/tlm}{https://github.com/DareenHarthi/tlm}}.
\end{abstract}


\section{Introduction}
\label{sec:intro}

Voice data are a treasure trove of information, revealing a diverse range of speaker characteristics, including their age
%. These can include demographic details such as geographical origin, race, or education level; medical traits like the use of medications, neurological, or genetic disorders; physical attributes including gender, height, weight, and facial structure; physiological traits such as age and blood pressure; and much more 
\cite{singh2019profiling}. 
%Given the depth and diversity of insights voice data can offer, the ability to model these vocal attributes accurately is becoming increasingly important. 

One widely explored task in this context, which we address in this paper, is age estimation from voice -- predicting a person's age from their speech recordings. Given the challenging nature of the problem, most approaches treat this as a classification task -- identifying the decadal age \textit{group} of the speaker \cite{kwasny2021explaining, sanchez2022age, badr2022voxceleb1}.  In this paper, we are, however, interested in the more challenging problem of deriving fine-grained estimates of the precise age of the speaker. This is generally formulated as a regression  where the goal is to predict a real-valued number (\textit{i.e.} the age) within a typical range, such as 10 to 80 years from features derived from the voice signal.  The relationship between the features and age is generally complex and must be learned from data comprising voice samples labelled with the age of the speaker. 

However, collecting age data for voice in this setting is challenging. It requires recording a person's voice and knowing their exact age at the time of recording. Consequently, age-labeled data with trustworthy age labels are limited, e.g. the TIMIT corpus \cite{garofolo1993darpa},
%; as far as we know, the most trusted dataset for age estimation is the TIMIT dataset \cite{garofolo1993darpa},  
with most other age-labelled datasets only including a guess of the speaker's age at the time of the recording. This limited availability of labeled data with a wide range of age information presents a significant challenge for learning the kind of complex relationship such as we may expect between voice features and age. Deep learning models, which excel at capturing such non-linear relationships, require large amounts of data for effective training. On the other hand, simpler models like linear regressions require far fewer data points to train, but often lack the flexibility to generalize well.

\begin{figure}[!t]
\centering
\includegraphics[width=0.7\columnwidth]{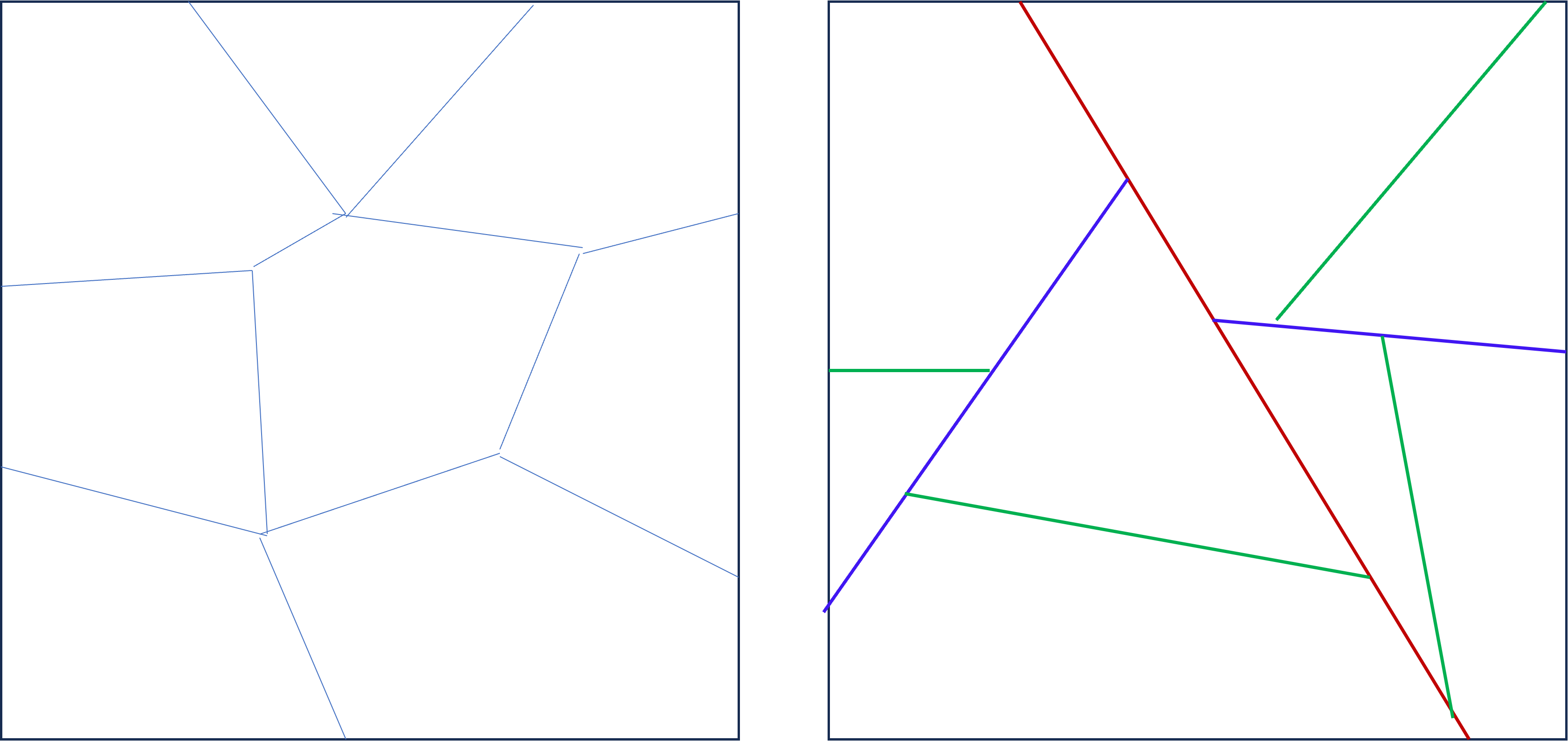}
\captionsetup{font=small}
\caption{(a) A convex tessellation of the input space. Ideally, both the tessellation and the linear estimator parameters within each cell must be optimized for prediction.  (b) Our hierarchical solution.  The space is recursively partitioned in a binary manner for locally optimal prediction. Here, the red line shows the first level partition, blue lines show the second level, and green lines show the third level. }\label{fig:tessellation}
%\vspace{-0.2cm}
\end{figure}

In this paper we propose an intermediate strategy, which aims to leverage the strengths of both approaches through a piece-wise linear function that retains part of the simplicity of low-order linear functions, while achieving the capacity of non-linear functions like neural networks. This model, which we call a \textit{Tessellated Linear Model} (TLM), operates by tessellating the space into convex cells that form a cover.  Within each cell the function is approximated by a linear function.  This is illustrated by Figure \ref{fig:tessellation}a. 

Optimizing the model, however, requires joint optimization of both, the tessellation and the local linear models, a combinatorial problem. 
%%% SHOULD I HAVE THIS HERE? 
%%Many existing methods in the optimization of piece-wise linear models simplify this joint optimization problem by making assumptions about the data space, often leading to suboptimal solutions \cite{hawkins1976point}. Our solution builds on the approach of Regression-via-Classification (RvC) \cite{torgo1997regression, memon2019neural, zhao2021hierarchical}. RvC techniques use classifiers to find change points that discretize the output into predefined categories, each representing a specific range of the continuous outcome.
%
To deal with this, we propose a hierarchical algorithm that finds a locally optimal tessellation through recursive binary partition of the space, as illustrated in Figure \ref{fig:tessellation}b. The approach also enables us to optimize the feature representation, derived using a deep neural network, jointly with the tessellated linear estimator.

%by learning to decompose the complex non-linear problem into multiple simpler linear functions -— a concept known as piecewise linear modeling. Piecewise linear modelling involves approximating a non-linear function using multiple linear models, each applicable to a specific region or interval of the input space. This approach requires the joint optimization of partition boundaries and local linear models, a combinatorial problem that is difficult to optimize. 

%Building upon this idea, we propose a new model called the Tessellated Linear Model. This model retains the simplicity of low-order functions, like linear functions, while achieving the capacity to approximate complex non-linear relationships akin to neural networks. The Tessellated Linear Model partitions the input space into convex sets that form a cover and models the function within each set using a linear model. Both the tessellation and the linear estimators within each region are optimized jointly for the overall prediction task.

We apply our model on the challenging voice biometric task of age prediction and demonstrate that it outperforms traditional neural models, outperforming all prior reported results on the TIMIT data by a significant margin, establishing a new SOTA on the task. 
%Our approach belongs to the RvC family and aims to learn an optimal piecewise linear predictor over linear tessellations of the input space.  We transform a complex combinatorial optimization problem into a more manageable learning problem by scanning optimal thresholds to partition the space based on the response variable. We then use binary classifiers to tessellate the space according to these chosen thresholds.

% The key contributions of this paper are:

% \begin{enumerate}
%     \item A simplified model for piecewise linear functions that uses linear regression and logistic regression classifiers to tessellate the space and fit a linear model in each segment.
%     \item A comparison of different training and inference strategies.
%     \item Evaluation on two regression tasks from voice data: age prediction and height prediction.
%     % \item An evaluation benchmark for age prediction based on Queen Elizabeth II's recordings from 1953 to 2021.
% \end{enumerate}

In the following sections, we present a brief review of both piece-wise linear models and age estimation from voice, formally describe the Tessellated Linear Model, present experimental results, and discuss the implications of this approach for regression problems in various domains.

\section{Related Work}
\label{sec:related_work}
Our work invokes two topics: the biometric prediction of age from voice, and piecewise linear models. We first present some of the relevant background on these topics.

\subsection{Age Estimation from Voice}
%A person's voice changes as they age \cite{singh2019profiling}. Knowing of this relationship, several researchers have attempted to estimate speakers' ages from voice recording. The approaches have generally fallen into two categories -- classifying the decadal age group of the speaker \cite{}, and estimating their actual age \cite{}. This paper addresses the latter problem.

Early approaches attempted to estimate age simple regressors such as support-vector regressors \cite{sadjadi2016speaker, bahari2012age},  models such as random forests \cite{singh2016short}, or statistical minimum mean squared error (MMSE) estimators based on statistical characterizations of the distributions of speech features \cite{ minematsu2003automatic}
. Speech features used conventional cepstral or Mel-frequency spectral or ceptral features \cite{ mahmoodi2011age, spiegl2009analyzing}, prosodic features such as pitch, energy, duration etc. \cite{li2013automatic}, features derived through signal-processing techniques that were expected to enhance speech characteristics related to age \cite{singh2016short, muller2007combining}, and those optimized for speech recognition. For instance, \cite{sadjadi2016speaker} used i-vectors with senone posteriors \cite{lei2014novel}, where neural networks trained on speech recognition tasks are used to estimate senones. This phonetically-aware approach resulted in a reduction of mean absolute error (MAE) from 5.0 to 4.7, highlighting the benefit of incorporating speech recognition features into age estimation models

%employed shorter analysis windows (1ms to 4ms) to focus on speech segments during glottal opening, in contrast to the conventional 25ms windows. These shorter windows were believed to better capture subglottal structures, potentially providing more accurate physiometric information from the voice. Additionally, research has explored the combination of short-term cepstral features, like Mel-Frequency Cepstral Coefficients (MFCCs), with utterance-based pitch information to enhance age estimation accuracy \cite{muller2007combining}. Other approaches, such as those proposed by \cite{li2013automatic}, incorporated prosodic features—such as pitch, energy, duration, and formants—alongside acoustic information to model age-related changes in speech.  

Recently, the field has shifted towards leveraging deep learning features for more effective age estimation \cite{kaushik2021end, kwasny2020joint, rajaa2021learning, gupta2022estimation}. Approaches have included both those that utilize self-supervised features such as Wav2Vec \cite{schneider2019wav2vec} and WavLM \cite{chen2022wavlm}, and features specialized for speaker identification or verification, such as x-vectors \cite{snyder2017deep, snyder2018x, wan2018generalized} or TitaNet \cite{koluguri2022titanet, tarashandeh2023agenet}.  These features have shown improvements in speaker identification due to the rich information they contain about speaker characteristics, and have been utilized by \cite{kwasny2021explaining, kwasny2020joint, ghahremani2018end} for age estimation. While these have resulted in improved age prediction, results have primarily been reported on decadal age categorization \cite{yucesoy2024speaker, kwasny2021gender, sanchez2022age, badr2022voxceleb1 }. Results reported on age \textit{regression}, \textit{i.e.} estimating the actual age of the speaker \cite{wang2024timit, gupta2022estimation, kaushik2021end, kwasny2020joint, kalluri2019deep }, are all worse than those we report.

%Approaches such as x-vectors \cite{snyder2017deep, snyder2018x, wan2018generalized}, originally developed for speaker verification and identification, along with features extracted from self-supervised models like Wav2Vec \cite{schneider2019wav2vec} and WavLM \cite{chen2022wavlm}, have gained prominence. These deep learning-based representations outperform traditional handcrafted features by learning directly from the entire speech signal, providing more robust and generalizable solutions. In our experiments, we employ TitaNet \cite{koluguri2022titanet} and WavLM features in conjunction with our proposed Tessellated Linear Model (TLM).

\subsection{Piece-wise Linear Models}

Piece-wise linear models  approximate complex functions by multiple  affine models, each valid within a specific partition or interval of the input space. When used to fit to the input-output relation of a given data, they require  joint optimization of both, the partition boundaries and the local linear models, a combinatorial problem. For univariate or multivariate time-series data, this requires the identification of the optimal ``knots'', or ``breakpoints'', where the model changes its linear behavior, and fitting separate affine functions between these breakpoints. Breakpoints may be identified through change-point detection algorithms, such as Davies' test \cite{davies1987hypothesis} and the Pruned Exact Linear Time (PELT) \cite{verzelen2023optimal} algorithm.

In the more generic 
%``MISO'' (multiple-input single output) setting where a relationship between a multi-variate input and a scalar output must be modelled,
multivariate setting, computing the optimal partition of the space is rather more complicated, being polynomial in the number of training points and exponential in the number of partitions \cite{jackson2010optimal}. The most common approach here is based on classification and regression trees \cite{loh2011classification}, which tessellate the space through hierarchical partitioning. The traditional CART model subsequently predicts different values in each partition, but the predictor in each partition is a constant, making it a piece-wise constant model, rather than a piece-wise linear model. Other regression-tree techniques such as \cite{friedman2006tree } do compute piece-wise linear models; however partition boundaries are constrained to be axis-aligned. Yet other methods such as \cite{raymaekers2023fast} permit more generic boundaries; however the optimization of these boundaries is typically performed using alternate statistical tests that are only proxies for the actual task -- accurate prediction of the dependent variable. An alternate solution that also uses a tree-based approach is Regression-via-Classification (RvC) \cite{torgo1997regression} which directly optimizes the partitioning for the prediction of the dependent variable. However, this class of methods too derives piece-wise \textit{constant} rather than piece-wise \textit{linear} functions.  In all cases, the techniques do not permit further optimization of the input \textit{features} themselves, over which the model is computed.

In this context it must be pointed out that neural networks with ReLU activations \cite{nair2010rectified} also model piece-wise linear functions, but unlike the other methods mentioned above are restricted to being continuous at partition boundaries. This implicit continuity constraint restricts model accuracy in low-data settings, where the number of partitions must also be low.
%but often require large amounts of data and strong regularization. Moreover, they assume the continuity of the piecewise linear function, which imposes strict boundaries between segments.

%The closest approaches to our proposed method are the segmented linear model with binary segmentation and the NRT algorithm. The former uses squared differences in the feature space to segment the data, while the latter employs a classifier with a set of thresholds (such as the median within each segment in NRT) to tessellate the space. However, to the best of our knowledge, no prior work has explored scanning for optimal thresholds based on the response variable and using the reduction in error at each level as a guide for tessellation—especially in the context of voice biometrics tasks.

Our proposed Tessellated Linear Model (TLM) fall generally under the regression tree family but places no constraints on the orientation or location of partition boundaries, and optimizes all components including the boundaries and linear models directly for the prediction of the dependent variable without any continuity restrictions at boundaries.
%, addresses this gap by partitioning the input space into convex regions and applying local linear regressors within each region. This approach allows for more flexible and accurate modeling of non-linear relationships, while maintaining the simplicity and interpretability of linear models.

% \begin{figure}[htbp]
%   \centerline{\includegraphics[width=\linewidth,height=0.2\textheight]{figures/tlm_syn.png}}

%   \caption{Left: A convex tessellation of the input space. An optimal TLM would determine both the tessellation and the linear estimator parameters within each cell, for optimal prediction. Right: Our hierarchical solution. The space is recursively partitioned in a binary manner for optimal tessellation. Here, the red line shows the first level partition, blue lines show the second level, and green lines show the third level.}
%   \label{fig}
% \end{figure}

\section{Tessellated Linear Model}
\label{sec:approach}
The  Tessellated Linear Model (TLM) partitions (or tessellates) the input space (of speech features) into a union of disjoint convex cells, and learns a separate linear model to predict the response variable (age in our case) in each cell. Both the tessellation, and the set of linear models are optimized for the prediction. Additionally, the underlying speech features over which the tessellation is defined may themselves also be optimized.

Let the input space $\mathcal{F} \subset \mathbb{R}^d$ and the response variable $y \in \mathbb{R}$. A convex tessellation  is a partition of $\mathcal{F}$ into $N$ disjoint convex cells:
\[
\tau(\mathcal{F}) = \{C_1, \dots, C_N\}
\]
satisfying $\bigcup_{n=1}^{N} C_n = \mathcal{F}$, $C_i \bigcap C_j = \varnothing$, and every $C_n$ is a convex subset of $\mathcal{F}$. Since the $C_n$ are convex and form a cover, they are all necessarily polytops \cite{gruber2007convex} bounded by hyperplanes, thus the tessellation can also be specified as $\tau(\mathcal{F}) = \mathcal{H}$, where $\mathcal{H}$ are the hyperplanes that separate the cells.

%As a convex cover, the cells of the tessellation are all necessarily polytopes [CITE] bounded by hyperplanes. Thus, the entire s

% In this section, we will describe the Tessellated Linear Model (TLM) for the regression task, which aims to partition the input space and fit separate linear models in each region. We optimize both the tessellation and the regression functions using a recursive binary splitting approach.

% Let the input space $\mathcal{X} \subset \mathbb{R}^d$ and the response variable $y \in \mathbb{R}$. The goal of the TLM is to partition $\mathcal{X}$ into convex cells where each cell can be modeled by a simple linear function. We define the tessellation $\tau$ as a partition of $\mathcal{X}$ into $N$ disjoint convex cells:
% \[
% \tau(\mathcal{X}) = \{C_1, \dots, C_N\}
% \]
% satisfying $\bigcup_{n=1}^{N} C_n = \mathcal{X}$, and each $C_n$ is a convex subset of $\mathcal{X}$. 

Within each cell $C_n$, the TLM fits a linear regression function with parameters ${\theta_{n}}$ ( $\theta_{n} = (r_{\theta_n}, b_n)$, where $r_{\theta_n}$ is the  regression coefficient vector and  $b_n$ is the bias) to approximate the response $y$. The linear model for an input $\mathbf{f} \in C_n$ is given by 
\[
\hat{y}_{\theta_n}(\mathbf{f}) = r_{\theta_{n}}^\top \mathbf{f} + b_n
\].

In our problem, the space $\mathcal{F}$ is itself a space of \textit{features} derived from speech signals, thus, corresponding to each input $\mathbf{x}$ is a feature $\mathbf{f}_\mathbf{x} \in \mathcal{F}$, such that $\mathbf{f}_\mathbf{x} = F(\mathbf{x}; \theta_F)$, where $F(.;\theta_F)$ is the \textit{feature extractor} with parameters $\theta_F$.

The overall TLM prediction for any input $\mathbf{x}$ is thus given by
\begin{equation}
   \hat{y}_{TLM}(\mathbf{x}) = \hat{y}_{\theta_{C_{\mathbf{f}_{\mathbf{x}}}}}
\label{eq:tlm} 
\end{equation}
where $C_{\mathbf{f}_{\mathbf{x}}}$ is the cell for $\mathbf{f}_\mathbf{x}$.

The TLM itself is learned to minimize the squared regression error. Representing $\Theta = \{\theta_n \forall C_n\}$, we can define the loss on the training set as
\[
\mathcal{L}(\mathcal{H}, \Theta, \theta_F) = \sum_{(\mathcal{x},y) \in \mathcal{D}} (y - \hat{y}(\mathbf{x}))^2
\]
Note that this loss is a function of both, the TLM parameters $\mathcal{H}$ and $\Theta$, and the feature extractor parameters $\theta_F$.
Optimizing the overall predictor thus requires minimizing $\mathcal{L}$ with respect to both the TLM parameters and the feature extractor parameters.

Simultaneous optimization of all three variables is difficult, so instead we resolve this through the following sequential steps:
\begin{align*}
    \hat{\mathcal{H}}, \hat{\Theta} & = \arg\min_{\mathcal{H}, \Theta} \mathcal{L}(\mathcal{H}, \Theta, \theta_F^0)\\
    \hat{\theta_F} & = \arg\min_{\theta_F} \mathcal{L}(\hat{\mathcal{H}}, \hat{\Theta}, \theta_F)
\end{align*}
where $\theta_F^0$ is a reasonable initial value of $\theta_F$. The first step initializes the feature extraction with a reasonable initial value $\theta_F^0$ and optimizes the TLM; the second freezes the TLM and optimizes the feature extraction. In principle the above two steps can be iterated, however we have found a single iteration to be sufficient. We describe the two steps below.

\subsection{Optimizing the TLM}
Optimizing the tessellation requires joint estimation of the set of all partitions $\mathcal{H}$ and the regression parameters $\Theta$. This is an infeasible combinatorial optimization problem \cite{grunbaum1969convex}.  Instead, we optimize it through recursive binary partitioning, where at each step we partition the data into two sets that are separated by a hyperplane, such that when a separate regression is applied to each side of the hyperplane, the overall regression error is minimized.  For even this simpler problem, we must propose candidate separating hyperplanes to select the best one from.  Our candidate set comprises the hyperplanes that best separate the data by the value of the response variable, namely the age. This leads to the following procedure that builds a tree of binary partitions, which can be shown to monotonically decrease prediction error at each step. 

%Let $\mathcal{Y}$ be the set of all response variable values in the current data set. For each $t \in mathcal{Y}$, let $H_t$ represent the hyperplane that best separates the data 

Starting with the root of the tree (containing the entire training data), at each node the tessellation 
%process $\tau$ 
is optimized recursively by scanning potential binary boundaries based on the response variable $y$. At each step, a linear classifier $h_t$ is trained to split the data into two regions: left ($y \leq t$) and right ($y > t$), where $t$ is the threshold that defines the boundary. The classifier $h_t$ assigns each point $\mathbf{f}_\mathbf{x}$, $(\mathbf{x}, y(\mathbf{x})) \in \mathcal{D}_n$ (where $\mathcal{D}_n$ is the training data at the current node) to either the left or right region:
\[
h_t(\mathbf{f}_\mathbf{x}) = 
\begin{cases} 
1 & \text{if } y(\mathbf{x}) \leq t \\
0 & \text{if } y(\mathbf{x}) > t
\end{cases}
\]

Once the data is split, separate linear models ${\theta_{\text{left}}}$ and ${\theta_{\text{right}}}$ are trained on the left and right subsets, respectively. The threshold $t$ is chosen to minimize the total prediction error after the split:
\begin{align*}
t^* = \arg\min_t \Bigg( 
    \sum_{\mathbf{x} \in \text{left}} (y - \hat{y}_{\text{left}}(\mathbf{x}))^2  + \sum_{\mathbf{x} \in \text{right}} (y - \hat{y}_{\text{right}}(\mathbf{x}))^2 \Bigg) 
\end{align*}

Initially, the parent node's regression model $r_{\theta_{\text{parent}}}$ is trained on the entire dataset.
The splitting process is repeated recursively, with new classifiers $h_\theta$ and regression models $\theta_n$ trained at each stage, until a stopping criterion is met. The recursion stops when a maximum depth $D$ is reached, or when no further meaningful splits can be made.

The final prediction for an input $\mathbf{x}$ is made by first computing features $\mathbf{f}_\mathbf{x}$ from it, and passing it through the series of classifiers in the hierarchical tree structure. At each node in the tree, a classifier $h_t$ guides $\mathbf{f}_\mathbf{x}$ to either the left or right child node, depending on the learned decision boundary at that level. This process is repeated recursively until $\mathbf{f}_\mathbf{x}$ reaches a leaf node, corresponding to a specific region $C_n$. Once $\mathbf{f}_\mathbf{x}$ is assigned to the correct region, the corresponding regression model ${\theta_n}$ is used to make the final prediction. Formally, the prediction is given by:
\[
\hat{y}(\mathbf{x}) = r_{\theta_{\tau(\mathbf{x})}}^\top \mathbf{x} + b_{\tau(\mathbf{x})}
\]
where $\tau(\mathbf{x})$ represents the tessellation procedure to reach node (or region) $C_n$, and $r_{\theta_{\tau(\mathbf{x})}}$ and $b_{\tau(\mathbf{x})}$ are the parameters of the linear model for that region.

The  procedure for training the TLM is shown in Algorithm \ref{alg:TLM}.

\begin{small}
\begin{algorithm}
\fontsize{8}{10}\selectfont

\caption{Tessellated Linear Model}
\label{alg:TLM}  % Label for reference
\begin{algorithmic}
\Procedure{BuildTree}{${D_n}$}
\State Initialize $t_n, \tau_n, \theta_{nl}, \theta_{nr}$
\State $\text{best\_reduction} \gets -\infty$
\State $\text{best\_threshold} \gets \text{None}$
\State $\theta_n \gets \text{Regressor}(D_n, \theta_n)$
\State $\text{error}_{np} \gets \text{RegressionError}(D_n, \theta_n)$
\For{$t_i \in t_n$}
    \State $(t_i^*, \tau_n) \gets \text{Classifier}(D_n, t_i, \tau_n)$
    \State $D_{nl}, D_{nr} \gets \text{Partition}(D_n, t_i^*)$
    \State $\theta_{nl} \gets \text{Regressor}(D_{nl}, \theta_{nl})$
    \State $\theta_{nr} \gets \text{Regressor}(D_{nr}, \theta_{nr})$
    \State $\text{error}_{nl} \gets \text{RegressionError}(D_{nl}, \theta_{nl})$
    \State $\text{error}_{nr} \gets \text{RegressionError}(D_{nr}, \theta_{nr})$
    \State $\text{total\_reduction} \gets \text{error}_{np} - \text{error}_{nl} + \text{error}_{nr}$
    \If{$\text{total\_reduction} > \text{best\_reduction}$}
        \State $\text{best\_reduction} \gets \text{total\_reduction}$
        \State $\text{best\_threshold} \gets t_i^*$
    \EndIf
\EndFor
\State $t_n^* \gets \text{best\_threshold}$
\State $D_{nl}, D_{nr} \gets \text{Partition}(D_n, t_n^*)$
\For{$D_n \in \{D_{nl}, D_{nr}\}$}
    \If{$D_n$ is pure}
        \State \textbf{continue}
    \Else
        \State \textsc{BuildTree}($D_n$)
    \EndIf
\EndFor
\EndProcedure
\end{algorithmic}
\end{algorithm} 
\end{small}

\subsection{Optimizing the features}
Once the TLM parameters are estimated, the predicted value for any input $\mathbf{x}$ can be computed as $\hat{y}_{TLM} (\mathbf{x}; \hat{\mathcal{H}}, \hat{\Theta}, \theta_F )$, where the parameters of the estimator are explicitly shown. More explicitly, this can be written as $\hat{y}_{TLM}(F(\mathbf{x}; \theta_F); \hat{\mathcal{H}}, \hat{\Theta})$, to indicate that this operates on features derived by the function $F(\mathbf{x}; \theta_F)$, typically a neural network in our setting. Since the linear classifiers and regressions on the TLM are all differentiable,  $\theta_F$ can be optimized to minimize $\mathcal{L} = \sum_{(\mathbf{x}, y) \in \mathcal{D}} (y - \hat{y}_{TLM}(F(\mathbf{x}; \theta_F); \hat{\mathcal{H}}, \hat{\Theta}))^2$ using backpropagation (and gradient descent).

\begin{table}[!ht]
\centering
\caption{A comparison between TLM and other age estimation models using TIMIT dataset for training and testing}
\resizebox{\linewidth}{!}{%
\begin{tabular}{lcc}
\hline \vspace{-0.2cm} \\

\textbf{Method} & \multicolumn{1}{l}{\textbf{Age RMSE}} & \multicolumn{1}{l}{\textbf{Age MAE}} \\
\hline  \vspace{-0.2cm} \\

Common sense test         &7.52                 & 5.55 \\
TLM with oracle         &1.18                 & 0.49 \\
\hline
Linear Regression& 5.79   & 4.49 \\ 

K-means and Linear Regression    &  7.69           & 5.88 \\ 
Random Forest & 6.58    & 4.85 \\ 

MLP with ReLU & 6.79   & 4.73 \\ 
% Linear Regression + mixup& 5.77  & 4.48 \\ 

% K-means and Linear Regression   + mixup &  6.61          & 5.22 \\ 

% Random Forest + mixup & 6.52    & 4.99 \\ 

% MLP with ReLU + mixup & -    & - \\ 
\hline
Singh et al. \cite{singh2016short}          & 8.16                    & 5.83 \\
NRT \cite{memon2019neural}       &9.13                   &  7.19 \\

Manav et al. \cite{kaushik2021end}(single-task)   & 7.16               & 5.03 \\

Manav et al. \cite{kaushik2021end} (multi-task)    & 8.07              & 5.63 \\

Gupta et al.\cite{gupta2022estimation} & 5.86    & 4.13 \\ 

\hline

TitaNet + TLM (Hard routing) &   5.62         & 4.09 \\ 

TitaNet + TLM (Soft routing)   &    5.49          & 4.02 \\ 

TitaNet + TLM  (Features Optimization)     & \textbf{5.36}             & \textbf{3.97} \\ 
\hline \hline

\end{tabular}
}
\label{table:age_prediction_comparison}
\end{table}
\section{Experimental Setup}
\label{sec:experiment}

\subsection{Data}
For our experiments, we utilized the TIMIT dataset to perform age prediction regression task. The TIMIT dataset consists of 4,610 samples—3,260 from male speakers and 1,350 from female speakers. The test set contains 1,680 samples—1,120 from male speakers and 560 from female speakers. The age distribution of the speakers ranges from 20 to 58 years.

\subsection{TLM }
\paragraph{Training Process}
We trained the Tessellated Linear Model (TLM) using 192-dimensional embeddings from the Titanet-Large model \cite{koluguri2022titanet}. We used Algorithm \ref{alg:TLM} to train the TLM. At each node of the tree, we scanned possible age thresholds and trained a logistic regression classifier (with a linear decision boundary) to obtain a binary split of the data. A modified version of mixup \cite{zhang2017mixup} that  mixes samples with similar age values was used to augment the data.
We selected the threshold and corresponding classification and regression models that minimized the total squared error.  This process was repeated recursively for each branch until the maximum tree depth was reached or further splitting was no longer meaningful. 

% \paragraph{Feature Optimization}
To optimize the features, after constructing the tree we froze it and optimized the Titanet embeddings with two residual blocks, each consisting of two fully connected layers with ReLU and Leaky ReLU activations (dropout: 0.2). The overall loss to optimize the features was the average of all classifiers (Binary Cross-Entropy) and regressors (Mean Squared Error) losses. The optimized features were used to predict age with the TLM.

% After constructing the tree, we optimized the Titanet embeddings using two residual blocks. Each residual block \cite{he2016deep} consisted of two fully connected layers with ReLU and Leaky ReLU activations (with a dropout probability of 0.2). The optimized features were then used to predict age using the TLM. The overall loss was calculated as the average of the classifiers and regressors losses. For the classifier, we used Binary Cross-Entropy (BCE) loss, and for the regression tasks, we used Mean Squared Error (MSE) loss.

\paragraph{Inference Strategies}
We designed two strategies for inference: \textbf{Hard Routing}: The classifier at each node routes the input to a leaf, where the associated linear regression model predicts the response value. \textbf{Soft Routing}: We use the linear regression models at every nodes through the path to predict the responses. At each node, the classifier’s probability determined the contribution of that node to the final prediction. The root node had a probability of 1, and for each subsequent node, if the sample was routed to the left, the contribution would be the product of the current node's left probability and the predicted value of the left child. The final result was computed as the summation of the predicted values weighted by their probabilities at each node.

% We use the linear regression models at every nodes through the path it to predict the response. Each node's classifier probability determines its contribution to the final prediction, with the final result being the weighted sum of predictions from all nodes.

% \paragraph{Baseline Models}

% We compared the performance of TLM with several baseline models:  2- models that have similar structures, including a MLP with ReLU activation, K-Means with Linear Regression (where we performed K-Means clustering on the data with $k = 10$, matching the number of regions in TLM, and trained a linear regression model for each cluster. The K-Means algorithm also obtains a Voronoi tessellation of the feature space), Random Forest, and simple Linear Regression. 3- In addition, we compared TLM against three models for age prediction models that was explained in section \ref{sec:related_work}. 

\paragraph{Baseline Models}
We compared the performance of TLM against several baseline models: (1) \textbf{Common Sense Test}, which predicts the mean of the data, and \textbf{TLM with Oracle}, where the response variable $y$ is assumed to be known. The tree is built based on $y$, and at test time, the input is routed using $y$, then we use the linear regression model in that leave for prediction. (2) Models models that have similar structures, including a MLP with ReLU activation, K-Means with Linear Regression (where we performed K-Means clustering on the data with $k = 10$, matching the number of regions in TLM, and trained a linear regression model for each cluster. The K-Means algorithm also obtains a Voronoi tessellation of the feature space), Random Forest, and simple Linear Regression. (3) Additionally, we compared TLM to three age prediction models that was explained in section \ref{sec:related_work}.

% : \cite{singh2016short}, which extracts spectral features using short windows (1ms to 4ms) and uses a random forest for prediction (holding the current state-of-the-art results for height prediction), \cite{kaushik2021end}, an end-to-end deep learning model that combines LSTM and attention mechanisms to predict the target variables, and \cite{gupta2022estimation}, which leverages wav2vec 2.0 embeddings and employs separate transformer encoders for male and female speakers to predict age and height.

\begin{figure}[htbp]
  \centerline{\includegraphics[width=1.0\linewidth]{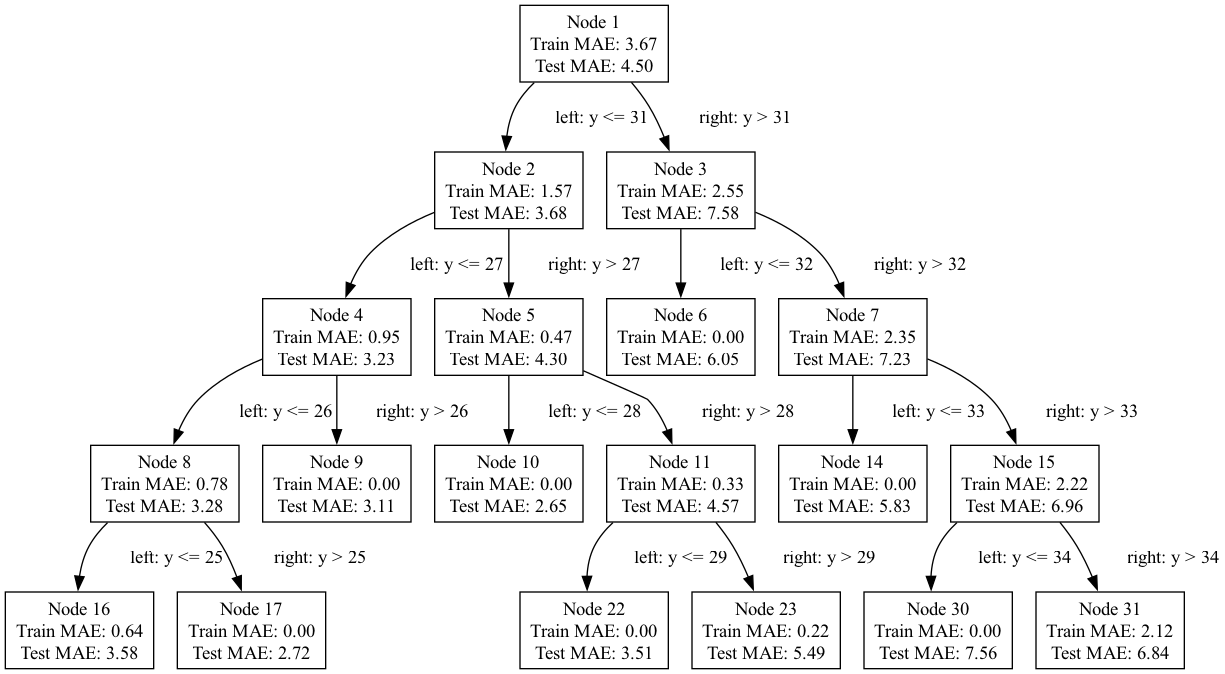}}

  \captionsetup{font=small}
  \caption{The tree shows the training and test MAE at each node of the TLM model. It illustrates the stepwise reduction in MAE with distinct thresholds defining the regions. Each parent node represents a binary decision based on a chosen threshold, and the MAE is evaluated only on the data samples routed to the region based on the threshold. It is evident that the error is lower for younger speakers, which corresponds to a higher number of training samples in these regions. }
  \label{fig:tlm_tree}
\end{figure}

% \begin{figure}[htbp]
%   \centerline{\includegraphics[width=\linewidth,height=0.3\textheight]{figures/tlm_real_depths.png}}

\begin{figure}[htbp]
  \centerline{\includegraphics[width=0.8\linewidth]{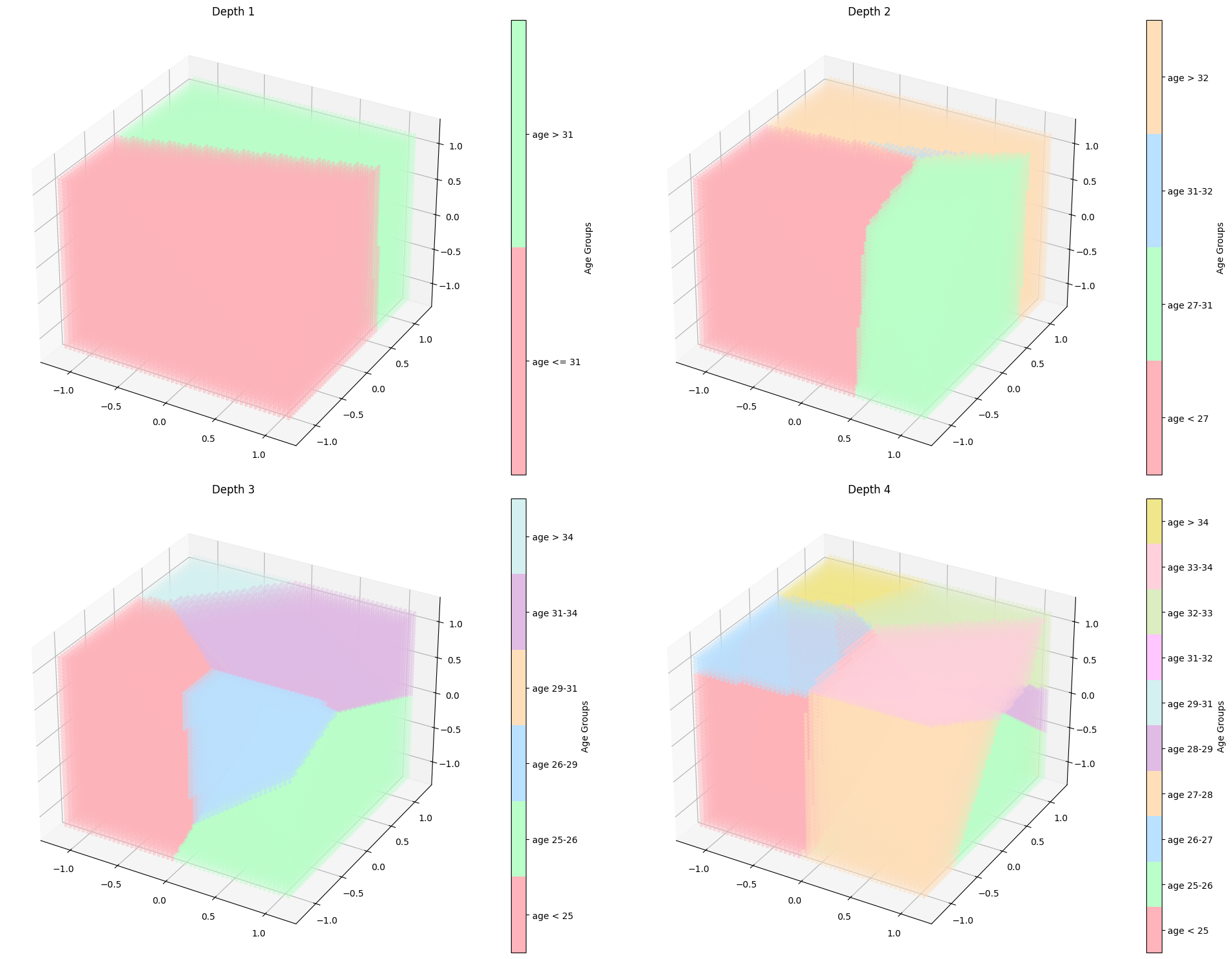}}
\captionsetup{font=small}
  \caption{The tessellation of the feature space is shown across the decision tree depths. Each region represents an age group, with colors indicating the predicted age. 
  %The regions are not always continuous; for example, the orange region (age 27-28) is adjacent to age groups less than 25, 25-26, and 31-32. 
  This segmentation shows how the model captures non-linear relationships between age and voice features in a piecewise-linear manner.}
  \label{fig:tlm_real_depths}
\end{figure}

\section{Results and Discussion}
\label{sec:results}
The Tessellated Linear Model (TLM) achieves a significant reduction in both MAE and RMSE compared to baseline methods. As shown in Table \ref{table:age_prediction_comparison}, TLM with feature optimization results in the lowest MAE of 3.97, the lowest error ever obtained on these data, by a significant margin. This indicates the effectiveness of dividing the input space into smaller regions, where simple linear models can operate effectively.  Interestingly, even without feature optimization, TLM with hard and soft routing still show competitive performance, with MAE values of 4.09 and 4.02, respectively, still signifianctly better than those obtained by complex deep learning models \cite{gupta2022estimation, kaushik2021end}.

The decision tree in Fig \ref{fig:tlm_tree} illustrates how the model achieves progressively lower errors at each depth, especially in regions with more training data, such as for younger speakers. Fig \ref{fig:tlm_real_depths} shows the tessellated space after training, with distinct regions for different age groups. Despite the smaller number of samples for older speakers, TLM still maintains competitive performance compared to the baseline models. Overall, the results suggest that TLM can generalize well across age groups by adjusting the partitioning of the feature space according to the data distribution.

\section{Conclusion}
\label{sec:conclusion}

We proposed the Tessellated Linear Model (TLM), a model that partitions the input space into convex sets and applies a linear model within each region. The tessellation and linear estimators are optimized jointly to learn an effective piecewise linear predictor. TLM uses binary classifiers to scan for optimal thresholds that partition the space based on the response variable.  The results show that TLM provides significant improvements over deep learning approaches, making it an efficient solution for structured prediction tasks like age estimation from voice data. Moreover, the piecewise linear nature of TLM enables better understanding of the model's decision-making process.

\bibliographystyle{ieeetr}  % Use the IEEE Transactions bibliography style
\bibliography{main}   % Replace 'references' with the name of your .bib file

\end{document}